\documentclass{article}
\usepackage{spconf,amsmath,graphicx,amsfonts, hyperref,siunitx}
\hypersetup{
    colorlinks=true,
    linkcolor=blue,
    filecolor=blue,      
    urlcolor=blue,
    citecolor=blue,
    pdftitle={Overleaf Example},
    pdfpagemode=FullScreen,
}

\title{Translated Skip Connections - Expanding the Receptive Fields of Fully Convolutional Neural Networks}
\name{J. Bruton, H. Wang}
\address{University of the Witwatersrand}
\begin{document}
%
\maketitle
\begin{abstract}
The effective receptive field of a fully convolutional neural network is an important consideration when designing an architecture, as it defines the portion of the input visible to each convolutional kernel. We propose a neural network module, extending traditional skip connections, called the translated skip connection. Translated skip connections geometrically increase the receptive field of an architecture with negligible impact on both the size of the parameter space and computational complexity. By embedding translated skip connections into a benchmark architecture, we demonstrate that our module matches or outperforms four other approaches to expanding the effective receptive fields of fully convolutional neural networks. We confirm this result across five contemporary image segmentation datasets from disparate domains, including the detection of COVID-19 infection, segmentation of aerial imagery, common object segmentation, and segmentation for self-driving cars.
\end{abstract}
\section{Introduction}
\label{sec:intro}
\footnote{Copyright 2022 IEEE. Published in 2022 IEEE International Conference on Image Processing (ICIP), scheduled for 16-19 October 2022 in Bordeaux, France. Personal use of this material is permitted. However, permission to reprint/republish this material for advertising or promotional purposes or for creating new collective works for resale or redistribution to servers or lists, or to reuse any copyrighted component of this work in other works, must be obtained from the IEEE. Contact: Manager, Copyrights and Permissions / IEEE Service Center / 445 Hoes Lane / P.O. Box 1331 / Piscataway, NJ 08855-1331, USA. Telephone: + Intl. 908-562-3966.}

In this paper, we propose an extension of the skip connection module that greatly increases the receptive fields of FCNs by taking advantage of their encoder-decoder structure. As opposed to a traditional skip connection, which skips an input from an earlier layer deeper into the network; we gradually translate the skipped inputs spatially further from their centre as they progress up the decoder. Since the decoder, composed of inverse convolutions, continually increases the resolution of the feature maps as we translate across the skipped input, the granularity of the translated receptive field continually decreases the further we move from the original kernel. Computational and memory burdens were an important consideration when developing the proposed module, and we have kept it in-line with other contemporary approaches to increasing receptive fields. We evaluate the proposed module on five different segmentation domains, all of which vary in resolution, data availability, and label complexity. The datasets range from medical image segmentation for COVID-19 infection to segmentation for self-driving cars. We aim to show that the increase in the receptive fields of FCNs provided by the proposed module has the potential to improve performance on a wide variety of image segmentation tasks.

Some background and an introduction to related work is required before we describe the proposed module, we provide this in section \ref{sec:background}. We then introduce the proposed module, the translated skip connection (TSC), in section \ref{sec:tsc}. In section \ref{sec:experiments}, we demonstrate the efficacy of the module empirically by embedding it in an architecture and comparing it to other architectures that increase the receptive field using strided pooling, dilated convolution, and strided convolution. We end the paper in section \ref{sec:conclusion} with concluding remarks and describe some of the future work that could build upon the results shown here.

\section{Background and Related Work}
\label{sec:background}

\subsection{Receptive Fields}\label{subsec:receptive_fields}

The ``receptive field" of a pixel or unit in a layer of the network is defined as the region of that layer's input that non-negligibly impacts the unit in question. This amounts to the region of the feature map covered by the kernel that provided that unit's value. If we propagate that concept back through an entire network, we can find the region of the original input to the network that affected the value of the unit under consideration. That region of the original input is called the effective receptive field (ERF), or just the ``effective field" \cite{luo2017erf}.

The ERF of the network often determines which tasks the network can undertake. Tasks that are more reliant on the shape of objects than their texture, or tasks that require the detection of very spatially dispersed objects, may be impossible without effective fields large enough to include the relevant features of the input space. There are a number of ways to increase the ERF of a network: (1) increasing the depth of the network, (2) using larger convolutional kernels, (3) resizing the input image, (4) using pooling, (5) using strided convolution or strided pooling, and (6) using dilated convolution.

\subsection{Expanding Receptive Fields}\label{subsec:dilated-convolution}

Increasing the depth of the network, resizing the input image, using larger convolutional kernels, and using pooling were all mentioned in section \ref{subsec:receptive_fields} as methods that increase the receptive fields of convolutional neural networks. All of these approaches have drawbacks; increasing the depth of neural networks often leads to vanishing gradients and over-fitting, besides a general increase in computational burden. Pooling and resizing the input image can dramatically decrease the dimensionality of the input data and can result in information loss, they also manipulate the data in a fixed way as opposed to learnable techniques which are more adaptive. While using larger convolutional kernels does increase the receptive field, it has been shown that larger convolutional kernels are disproportionately computationally expensive; architectures with more layers and smaller kernel sizes are more efficient despite having the same number of parameters \cite{szegedy2016inception}. Strided convolution provides a learnable way to increase the ERF of an FCN however it typically replaces strided pooling, which means an increased number of parameters and a greater propensity to overfit.

Dilated convolution \cite{holschneider1990dilate} provides a way to maintain the kernel size while increasing the receptive field of convolutional kernels in a memory efficient way. We also define the \textit{dilation rate} as the horizontal, vertical, or diagonal distance from each weight to its nearest neighbours in the filter. Traditional convolution is said to have a dilation rate of \(1\). As shown by \cite{yu2015multi} and \cite{oord2016wave}, dilated convolution out-performs traditional convolution in a variety of cases.

While dilated convolution does not require an increase in the number of parameters used in each kernel, and so has similar memory requirements to traditional convolution, it does create gaps in the kernel. These gaps mean that the optimisations generally applied to matrix multiplication are not effective when dilated convolution is being used; because of this, dilated convolution often takes up to three times longer than traditional convolution despite not using more memory.


\subsection{Skip Connections} \label{subsec:skip}

Skip connections \cite{he2016skip} serve to counteract the problem of deeper networks increasing training error through continually vanishing gradients. By connecting layers in the network directly to the inputs of layers that occurred earlier in the network, we curtail the tendency for weights and biases to degrade the inputs of deeper layers. As weights deeper in the network get smaller, these residual blocks may easily tend towards the identity function instead of tending to an output of zero, particularly when the ReLU activation function is used. Thus, skip connections allow us to add layers to CNNs almost indiscriminately; if the layers are of no benefit then the output of the network will not be altered.

We consider two different implementations of skip connections, additive and concatenative. The first, \(\textbf{f(x)}+\textbf{x}\), we refer to as an ``additive skip connection". This is simply the element wise summation of the output of one layer with the input of an earlier layer in the network. The latter ``concatenative skip connections", denoted \(\textbf{f(x)}\oplus\textbf{x}\), concatenate the two layers together, and thus requires extra parameters to process.

\section{Translated Skip Connections} \label{sec:tsc}

We propose translated skip connections (TSCs), a novel redesign of skip connections that greatly increases the effective field of FCNs while only marginally increasing the number of parameters in the network. The proposed module also includes a traditional additive skip connection, thus still avoiding the problem of vanishing gradients.

\begin{figure}
    \centering
    \includegraphics[width=8cm]{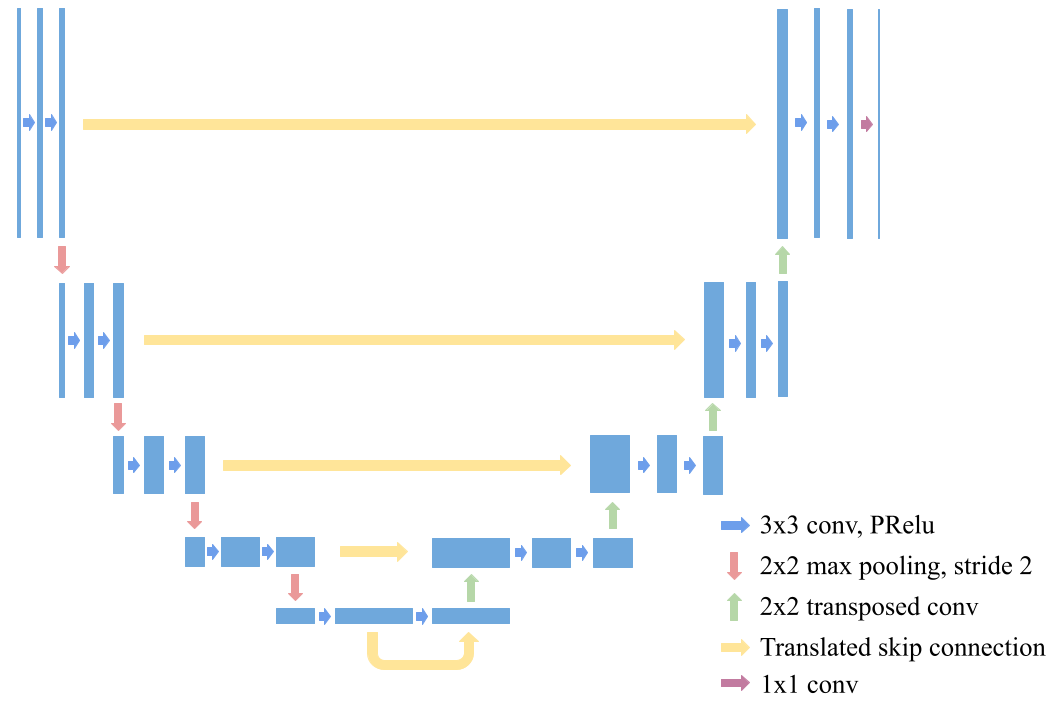}
    \caption{Diagram of the proposed TSC architecture}
    \label{fig:tsc_architecture}
\end{figure}

We define the translation function \(\textbf{T}:\mathbb{R}^{m\times n}\to\mathbb{R}^{m\times n}\) that shifts an input matrix in a given direction by a given factor. For example, given a feature map \(\textbf{X}\), \(\textbf{T}_{\uparrow}(\textbf{X}, \frac{1}{4})\) would translate the feature map upwards one fourth the height of the feature map, with the top of the feature map reentering at the bottom.

The TSC architecture used in our experiments is depicted in Fig. \ref{fig:tsc_architecture}. The architecture combines ideas from two earlier networks, called V-Net \cite{milletari2016vnet} and U-Net \cite{ronneberger2015unet}; we will refer to this architecture as TSC-Net. Each skip connection - represented by the yellow arrows - includes a single additive skip connection and three translated concatenative skip connections that translate the encoder output left, up, and diagonally, respectively. The translation factor in each TSC is \(\frac{l}{\textbf{D}+1}\), where \(l\) is the integer index of the layer in the network, from \(1\) to \(5\) starting at the bottleneck, and \(\textbf{D}\) is the network's depth, \(5\). Given the notation defined above, each of the five skip connections in the network can be described with the following equation, where \textbf{X} is the skipped input to the layer, l is the index of the layer, and \(\textbf{f}(\textbf{X})\) is the output of the previous layer:
\begin{equation}
(\textbf{f}(\textbf{X})+\textbf{X}) \oplus \textbf{T}_{\leftarrow}(\textbf{X}, \frac{l}{6}) \oplus \textbf{T}_{\uparrow}(\textbf{X}, \frac{l}{6}) \oplus \textbf{T}_{\nwarrow}(\textbf{X}, \frac{l}{6})
\end{equation}

This equation comprises a single TSC, so each TSC is made of a single additive skip connection, \(\textbf{f}(\textbf{X})+\textbf{X}\), and three concatenative skip connections, denoted \(\oplus\).

The translational equivariance demonstrated by CNNs is typically considered advantageous but it poses a problem for the proposed module. To ameliorate this, we implement a variant of CoordConv \cite{liu2018intriguing} by concatenating the coordinate of each pixel to its feature vector and normalising it. The transformation we apply to each pixel's feature vector is:
\begin{equation}
(R, G, B)\rightarrow{(R, G, B, \frac{x}{W}, \frac{y}{H})}
\end{equation}
yielding the five input channels of the network. Rather than translation equivariance being inevitable, this modification makes it optional; we show the benefit of this modification in the next section.

\section{Experiments and Results} \label{sec:experiments}

\subsection{Methodology}\label{subsec:methodology}

The five architectures we consider are constructed to compare TSCs to the other approaches of increasing the receptive field introduced in section \ref{subsec:receptive_fields}. Our implementations of these architectures, using Pytorch with Pytorch Lightning, are available on GitHub under a GNU General Public License\footnote{\url{https://github.com/JoshuaDBruton/TSC}}.

The baseline model that we consider is the popular U-Net architecture \cite{ronneberger2015unet}. U-Net uses \(2\times2\) strided max pooling to scale down the input image and thus increase the effective field. We extend the U-Net architecture twice, using dilated convolution with dilation rates of both two and three, as dilation rates higher than three resulted in poor performance. The fourth architecture we examine is a scaled down version of V-Net, which we refer to as B-Net; the primary difference is replacing U-Net's max-pooling with strided convolution.

We compare the four architectures discussed above to the TSC architecture, TSC-Net, depicted in Fig. \ref{fig:tsc_architecture}. TSC-Net would have more parameters than U-Net and B-Net, because of the two extra concatenative skip connections in each decoder layer. To prove that it is the placement of TSC-Net's parameters and not their quantity that improves performance, we reduce the number of filters so that it totals approximately \(9.6\) million parameters, fewer than B-Net.

All five architectures are trained on five different datasets for 100 epochs; they are: Landcover.ai \cite{boguszewski2020landcover}, Inria Image Labelling \cite{maggiori2017inria}, Carla \cite{dosovitskiy2017carla}, the COVID-19 CT scans dataset \cite{ma_jun_2020_3757476, Glick2020Radio}, and PASCAL VOC2012 \cite{everingham2010pascal}. We measure the performance of all the architectures with the Mean Intersection Over Union Metric \cite{rezatofighi2019miou}, and report the average of the maximum validation MiOU attained by each of the models.

\subsection{Results}\label{subsec:results}

\begin{figure*}
    \centering
    \includegraphics[width=10cm]{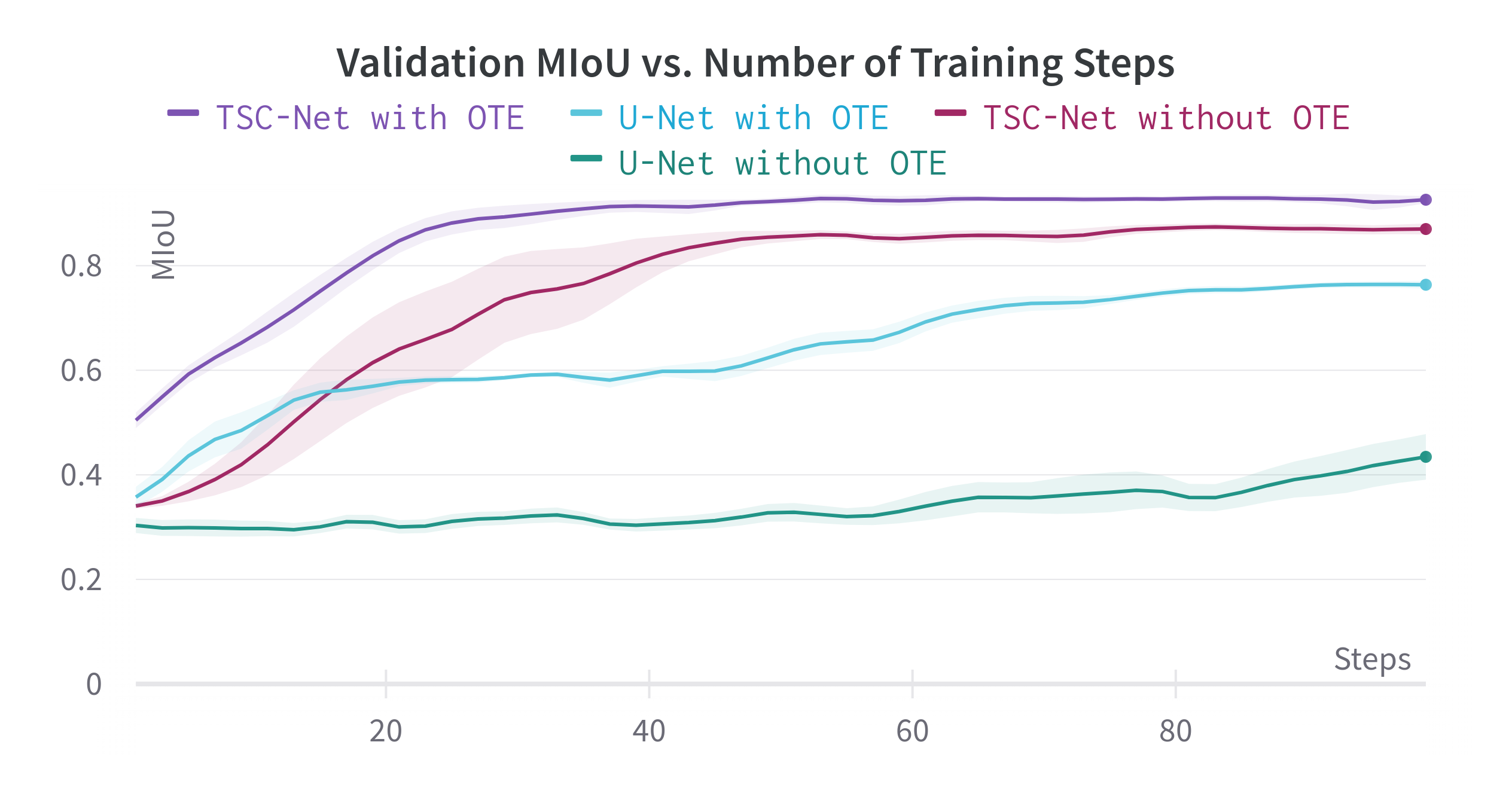}
    \caption[TSC Ablation Study]{Chart depicts the validation MIoU of four different models while training on the top and left rectangle dataset, each line is averaged over five training regimes, with the standard error depicted in a lighter shade}
    \label{fig:tsc_ablation_chart}
\end{figure*}

We must first establish whether TSC satisfies its intended purpose, that is, geometrically expanding the receptive fields of FCNs. We also test whether a network, after having its receptive field expanded by TSC, benefits from optional translational equivariance (OTE). This experiment is conducted on a synthetic texture analysis dataset available online \footnote{\url{https://webwtad.firebaseapp.com/}}. In this experiment, we make use of two FCN architectures, U-Net and TSC-Net. We train these networks with and without OTE on the synthetic dataset for 50 epochs each, averaging the results for each network over five training runs. The results of this experiment are shown in figure \ref{fig:tsc_ablation_chart}. As can be seen there, every addition of OTE and TSC results in a performance improvement compared to the baseline model; but, more interesting conclusions can also be drawn from this result.

Firstly, U-Net without OTE's performance barely improves throughout training, and - contrary to the norm - the stability of the model decreases the more it trains, signifying its inability to solve this task. The rest of the architectures all show convergence; however, the least stable architecture is TSC-Net without OTE. We argue that this is because the information provided by TSC is quite complex for the model to interpret without OTE. The model does not know where in the image the kernels in the TSC portion of its feature map are coming from. Two other results attest to this argument: when we add OTE to TSC-Net, the stability of convergence notably improves, and, at the beginning of training, U-Net with OTE performs better than TSC-Net without OTE until TSC-Net learns to make use of the complex information the TSC module provides.

\begin{table*}[h]
\caption{Comparison of baseline and dilated convolution with TSC}
\begin{center}
\begin{tabular}{l c c c c c }
\hline
\multicolumn{6}{c}{\textbf{Average of Maximum Mean Intersection Over Union$^{\mathrm{a}}$$^{\mathrm{b}}$}} \\
\cline{1-6} 
\textbf{Dataset} & \textbf{U-Net} & \textbf{Dilation 2} & \textbf{Dilation 3} & \textbf{B-Net} & \textbf{TSC-Net} \\
\hline
COVID-19 & 0.86 \(\pm\) 8\text{e}$^{-3}$ & 0.87 \(\pm\) 7\text{e}$^{-3}$ & 0.86 \(\pm\) 5\text{e}$^{-3}$ & 0.89 \(\pm\) 2\text{e}$^{-3}$ & \textbf{0.90} \(\pm\) 1\text{e}$^{-4}$\\
\hline
IILD & 0.72 \(\pm\) 0.01 & 0.74 \(\pm\) 3\text{e}$^{-3}$ & 0.75 \(\pm\) 4\text{e}$^{-3}$ & 0.79 \(\pm\) 2\text{e}$^{-3}$ & \textbf{0.81} \(\pm\) 1\text{e}$^{-4}$ \\
\hline
Landcover.ai & 0.45 \(\pm\) 0.01 & 0.49 \(\pm\) 6\text{e}$^{-3}$ & 0.50 \(\pm\) 9\text{e}$^{-3}$ & 0.50 \(\pm\) 5\text{e}$^{-3}$ & \textbf{0.53} \(\pm\) 6\text{e}$^{-3}$ \\
\hline
Carla & 0.73 \(\pm\) 6\text{e}$^{-3}$ & 0.71 \(\pm\) 9\text{e}$^{-3}$ & 0.70 \(\pm\) 2\text{e}$^{-3}$ & \textbf{0.76} \(\pm\) 3\text{e}$^{-3}$ & \textbf{0.76} \(\pm\) 2\text{e}$^{-3}$ \\
\hline
VOC20212 & 0.48 \(\pm\) 0.01 & 0.49 \(\pm\) 0.01 & \textbf{0.50} \(\pm\) 4\text{e}$^{-3}$ & 0.44 \(\pm\) 1\text{e}$^{-3}$ & \textbf{0.50} \(\pm\) 4\text{e}$^{-3}$ \\
\hline
\multicolumn{6}{l}{$^{\mathrm{a}}$Ranges from 0.0 to 1.0, closer to 1.0 is better} \\
\multicolumn{6}{l}{$^{\mathrm{b}}$Averaged over 3 trains for IILD and Landcover.ai, and 5 trains for the remaining datasets}
\end{tabular}
\label{table:main_results}
\end{center}
\end{table*}

Table \ref{table:main_results} contains experimental results for all five of the architectures we considered on all five datasets. All architectures achieve their greatest MIoU on the COVID-19 dataset, which is expected given that U-Net was initially designed for application to medical imaging and the other architectures are extensions of U-Net. B-Net's strided convolution provides learnable down-sampling and thus greater expressiveness, and in this domain did not cause over-fitting because the training and testing sets are fairly consistent. Despite not having strided convolution, TSC-Net's translated skip connections allow for a similar increase in expressiveness but also provide a larger ERF, and so TSC-Net slightly outperforms B-Net.

Following the COVID-19 dataset, we present results from IILD and Landcover.ai, two satellite imaging datasets with similar characteristics; unlike the COVID-19 dataset, satellite image segmentation certainly requires large receptive fields as objects typically span wide areas and their class is often inferred from the surrounding region. IILD and Landcover.ai are also large datasets, and so over-fitting is not a concern. This explains why larger, more learnable, receptive fields consistently improve performance on both datasets. Landcover.ai has a greater label-complexity, hence the lower MIoUs achieved by all architectures on that dataset, however the relative performance of the architectures was markedly similar in both domains.

The next result in Table \ref{table:main_results} is for the CARLA segmentation for self-driving cars dataset. It is a synthetic dataset, and so the textures and the gradients between them are simplistic compared to real-world images; this necessarily means that higher receptive fields will confer less benefit. Thus, the expressivity of each architecture will have more impact on their performance. B-Net's learnable down-scaling and TSC-Net's learnable skip connections are the source of their extra expressivity and better performance.

Finally, we present the results for the PASCAL VOC2012 dataset; the most complex domain we consider, it contains 22 labels, so random performance would result in an MIoU below 0.05. VOC2012 is also a small dataset, with approximately 1,500 training images, adding to the difficulty. Typically, researchers would apply deeper architectures to solve this domain, but since our primary concern is relative performance, our models are sufficient. For the first time, B-Net achieved the worst performance of all the architectures, this is likely due to overfitting. Only two of the architectures achieved the satisfactory margin of 0.5, the dilated model with dilation rate 3, and TSC-Net. Unsurprisingly, these are the architectures with the largest, learnable receptive fields; these receptive fields are an important quality on this dataset due to the large amount of correlated classes, for example, horse and grass or desk and monitor. Both of these architectures also make use of strided pooling when down-sampling, this counteracts the overfitting that befell B-Net's strided convolution.

\section{Conclusion}\label{sec:conclusion}

In this paper, we proposed a module for expanding the receptive fields of FCNs. We also investigated alternative approaches to increasing the receptive fields of FCNs. We considered dilated convolution, strided pooling, strided convolution, and, our proposed module, TSCs. Our experimental results, acquired across five disparate image segmentation datasets, show that using TSCs led to competitive or superior performance when compared to the other methods considered in this paper.

Future work should consider alternative implementations of TSCs that could extend to traditional neural networks. Alternatively, conducting more experiments with TSCs embedded in larger architectures that combine the techniques compared in this manuscript would also be beneficial. Finally, our results suggest that including strided pooling and strided convolution in an alternating fashion in the encoder layer of an FCN could simultaneously improve expressivity and curb over-fitting, which could provide for an interesting finding.

\bibliographystyle{IEEEbib}
\bibliography{refs}

\end{document}